\documentclass[5p]{elsarticle}

\usepackage{natbib}
\bibpunct{(}{)}{;}{a}{,}{,}
\usepackage{amsmath}

\usepackage{mathptmx}       
\usepackage{helvet}         
\usepackage{courier}        
\usepackage{type1cm}        
\usepackage{graphicx}        
\usepackage{multicol}        
\usepackage{multirow}
\usepackage[bottom]{footmisc}

\usepackage[caption=false,font=footnotesize]{subfig}

 \usepackage[usenames,dvipsnames]{pstricks}
 \usepackage{epsfig}
 \graphicspath{{images/}}
 \usepackage{pst-grad} 
 \usepackage{pst-plot} 
 
 \usepackage[nolists]{endfloat}

\usepackage{color}

\newcommand{\me}[1]{\mbox{\emph{#1}}}

\begin{document}

\begin{frontmatter}

\title{Modelling Cyber-Security Experts' Decision Making Processes\\using Aggregation Operators}
\author{Simon Miller}
\ead{s.miller@nottingham.ac.uk}
\author{Christian Wagner}
\ead{christian.wagner@nottingham.ac.uk}
\author{Uwe Aickelin}
\ead{uwe.aickelin@nottingham.ac.uk}
\author{Jonathan M. Garibaldi}
\ead{jon.garibaldi@nottingham.ac.uk}
\address{Intelligent Modelling and Analysis (IMA) Research Group / Lab for Uncertainty in Data and Decision Making (LUCID) / Horizon Digital Economy Research\\School of Computer Science, University of Nottingham, Nottingham, NG8 1BB, UK.}

%

\begin{abstract}

An important role carried out by cyber-security experts is the assessment of proposed computer systems, during their design stage. This task is fraught with difficulties and uncertainty, making the knowledge provided by human experts essential for successful assessment. Today, the increasing number of progressively complex systems has led to an urgent need to produce tools that support the expert-led process of system-security assessment. In this research, we use Weighted Averages (WAs) and Ordered Weighted Averages (OWAs) with Evolutionary Algorithms (EAs) to create aggregation operators that model parts of the assessment process. We show how individual overall ratings for security components can be produced from ratings of their characteristics, and how these individual overall ratings can be aggregated to produce overall rankings of potential attacks on a system. As well as the identification of salient attacks and weak points in a prospective system, the proposed method also highlights which factors and security components contribute most to a component's difficulty and attack ranking respectively. A real world scenario is used in which experts were asked to rank a set of technical attacks, and to answer a series of questions about the security components that are the subject of the attacks. The work shows how finding good aggregation operators, and identifying important components and factors of a cyber-security problem can be automated. The resulting operators have the potential for use as decision aids for systems designers and cyber-security experts, increasing the amount of assessment that can be achieved with the limited resources available.

\end{abstract}


\end{frontmatter}

\section{Introduction}
\label{sec:Intro}

As Internet use becomes ever more pervasive in day-to-day life for tasks including internet banking, e-commerce and e-government, the risk of cyber-crime is a growing concern (see \cite{Detica2011}, \cite{Anderson2013} and \cite{McAfee2014}). Assessing security risks associated with proposed systems in their design phase is a non-trivial task that involves managing multiple sources of uncertain information. For example, it is very difficult to estimate the costs of a successful attack, the likelihood of a rare attack, as the tools and technologies available to attackers/defenders are constantly changing \citep{Tregear2001}. Because of these difficulties, a great deal of expertise is required to carry out such assessments. Typically, cyber-security experts are employed as they can provide comprehensive assessments based on considerable expertise and insight while also being able to assess the viability of existing cyber-security tools and processes (e.g., anti-malware software). From a computational point of view, their work can be likened to that of a highly complex aggregation function, considering a large number of uncertain data sources (e.g., other experts, users, systems designers and security software / hardware) and fusing these sources to build overall security assessments. However, there is often a shortage of the level of cyber-security expertise required to carry out detailed assessments, leading to an urgent requirement for techniques and tools that can be used to support experts, reducing their workload, and systems designers, using models of expert knowledge to obtain estimates of security for system designs.

As stated previously, the job of performing security assessments is comparable to a complex aggregation function, fusing multiple sources of disparate data to form an overall assessment. In order to replicate this process in a computational model, aggregation operators such as the arithmetic mean, Wei\-ght\-ed Average (WA), and Ordered Weighted Average (OWA) could be considered. As will be shown later, WAs and OWAs allow the application of weightings to specific objects and specific positions in an ordering, respectively. We will employ both of these methods in this research, exploiting their characteristics to produce fused assessments. WAs are used to compute assessments of security components using sub-assessments of their characteristics, and OWA operators are used to compute salience/difficulty rankings of specific technical attacks using security component assessments. A difficulty when using WAs and OWAs is finding suitable weightings for a particular task, as there are an near-infinite number of possibilities. Evolutionary Algorithms (EAs) (\cite{Holland1975} and \cite{Goldberg1989}) have been shown to be useful in tasks involving a large search area, including searching for OWA weights \citep{Nettleton2001}. In the research presented in this paper, we will employ EAs to search for appropriate weights for WAs and OWAs for use in a security assessment problem.


The data set used is from a decision making exercise that was conducted at GCHQ, Cheltenham, UK, the UK government's signals intelligence and information assurance agency. Thirty nine GCHQ selected cyber-security experts including system and software architects, security consultants, penetration testers, vulnerability researchers and specialist systems evaluators, took part in two survey exercises. In the first, they were asked to rank a set of ten technical attacks in order of how difficult they would be to complete without being noticed. The set included attacks via a Voice Over IP (VOIP) client, a malformed document via email and a malicious website. In the second exercise, experts were asked to rate the difficulty of compromising / bypassing the twenty six security components that make up the ten attacks from the previous exercise. Security components included anti-virus software, cryptographic devices and firewalls. They were also asked to rate specific characteristics of each component, for example, the \emph{complexity} of a component or the \emph{public availability of tools} that could be used by a potential attacker to compromise / bypass the component. Undertaking this type of survey can identify particularly weak points in a system, and thus potential `breach-points' in the system. As such, this activity is an important part of the security assessment process. The result is a data set containing three levels of assessment:

\begin{figure}[tb]
\centerline{\includegraphics[width=5cm]{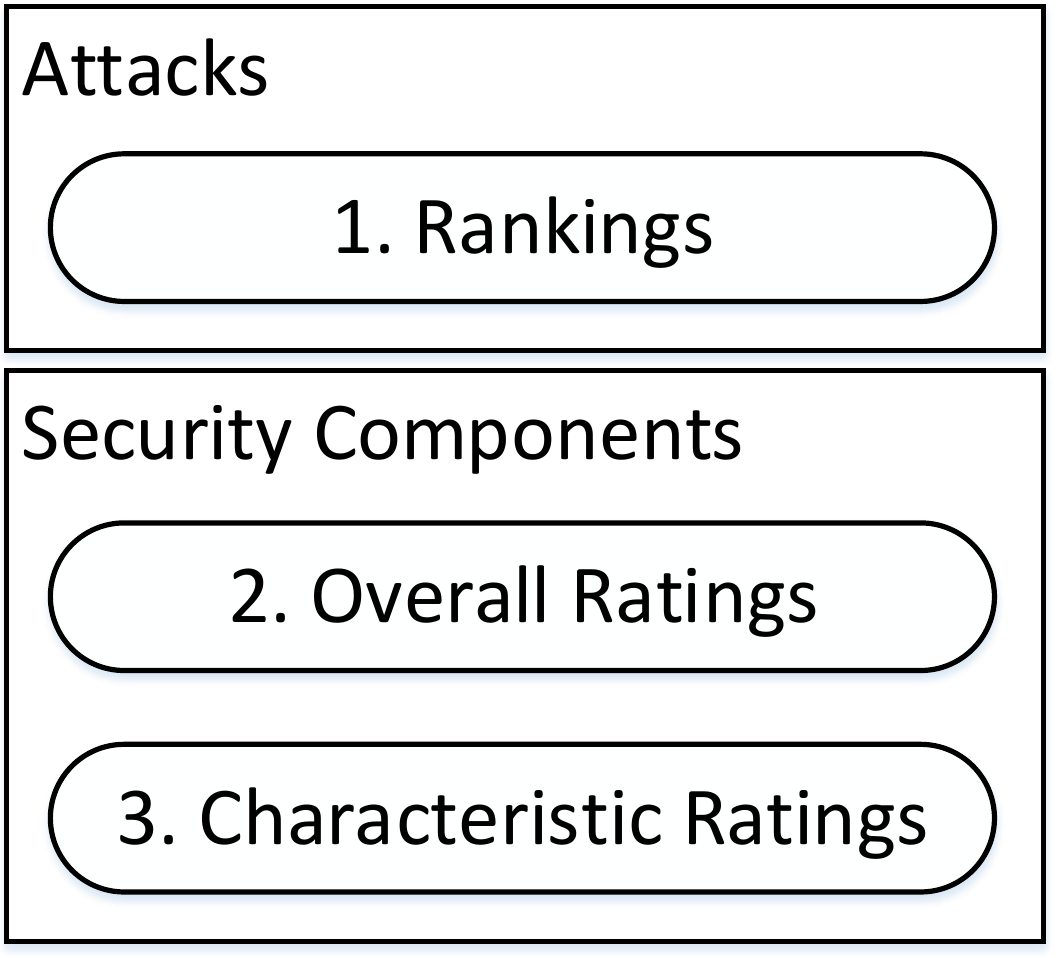}}
\caption{Structure of Assessment Data}
\label{fig:levels}
\end{figure}

\begin{enumerate}
	\item Rankings of technical attacks on a proposed system.
	\item Ratings of the level of security offered by security components in the proposed system.
	\item Ratings of specific characteristics of security components in the proposed system.
\end{enumerate}

Figure~\ref{fig:levels} shows the structure of the data set. With this data set we will show how, using WAs and OWAs with EAs, the following can be achieved:

\begin{enumerate}
	\item Ratings of the difficulty of attacking and evading security components using ratings of specific characteristics.
	\item Indication of the relative contribution each characteristic makes to the overall difficulty rating for a security component.
	\item Rankings of specific technical attacks using ratings of security components.
	\item Indication of the relative contribution components make to the attack ranking based on their difficulty rating.
	\item A combination of the previous aggregations, in which rankings of specific technical attacks are computed using security component ratings that have been created using ratings of specific characteristics.
\end{enumerate}

As we have data at all three levels, we are able to compare the derived security component ratings and attack rankings with experts' actual ratings / rankings to validate them. Potentially, this method could be used with a database of existing ratings for components / characteristics to aid in security assessments on proposed systems, reducing load on the cyber-security experts who currently carry out such assessments.

The paper is structured as follows. 
Section~\ref{sec:Back} provides an overview of the problem of performing security assessments on information systems, a review of machine learning approaches to this problem, WA/OWA operators and EAs. 
Section~\ref{sec:DE} describes the decision making exercise conducted with cyber-se\-cu\-rity experts at GCHQ, while
Section~\ref{sec:imp} details how EAs, OWAs and WAs have been implemented for this study. 
In Section~\ref{sec:Comp}, overall difficulty ratings for security components are produced using ratings of specific characteristics of each component. Moreover, we also produce weightings that signify the contribution each characteristic makes to the overall difficulty rating. 
Then, in Section~\ref{sec:Attack} rankings for specific technical attacks are produced using difficulty ratings of their constituent components, along with weightings that give an indication of which components contribute most to the rankings. 
Section~\ref{sec:All3} combines the processes of the previous two sections to derive ratings and rankings for specific technical attacks from ratings of specific characteristics of their constituent components. Finally, the outcomes of the work are discussed in Section \ref{sec:Disc}, and Section \ref{sec:Conc} summarises the contributions of the research and considers future directions of research.

\section{Background}
\label{sec:Back}


\subsection{Cyber-Security Assessment}
\label{sec:riskassess}

In current times, the use of the Internet and cellular networks has become the norm for a variety of every-day tasks including banking, e-mail, e-government, e-commerce, VOIP, social networking and mobile telecommunications. An ever-increasing number of sensitive interactions are taking place online, making them susceptible to attack by cyber-criminals who may read, modify or delete sensitive data. Concurrently, the volume and availability of tools to aid would-be attackers in their acts are also increasing. These tools lower the barrier to entry for attackers, as far less technical expertise is required to operate them than to launch an attack from scratch. Exacerbating the situation further is the fact that it is often unclear exactly how defensive security products achieve their claims, and indeed, whether they are effective at all.

With the increasing threat of cyber-crime the practice of carrying out security assessments on proposed systems has become a critical part of the systems design process. Understanding the threats that attackers pose to a proposed system, their consequences, how those threats can be addressed and what represents an appropriate level of security is a complex problem involving many factors. These factors include \citep{Tregear2001}:

\begin{enumerate}
	\item Data is limited on certain threats, such as the likelihood of a sophisticated hacker attack and the costs of damage, loss, or disruption caused by events that exploit security weaknesses.
	\item Some costs, such as loss of customer confidence or disclosure of sensitive information, are inherently difficult to quantify.
	\item Although the cost of the hardware and software needed to strengthen controls may be known, it is often not possible to precisely estimate the related indirect costs, such as the possible loss of productivity that may result when new controls are implemented.
	\item Even if precise information were available, it would soon be out of date due to fast paced changes in technology and factors such as improvements in tools available to would-be intruders.
\end{enumerate}

The consequence of these difficulties is that it is necessary for highly experienced cyber-security experts to carry out assessments of system designs including their hardware/devices, architecture, security software and practices before they are implemented. The goal of such assessments is to ensure that there is a proportionate level of security offered by the proposed system in line with the consequences of a successful attack. Clearly, investing too little in security measures leaves a system exposed to an unacceptable level of risk, however, investing too much is also a problem as it involves expending an unnecessary amount of money and effort on security when the consequences of a successful attack are considered.

Unfortunately, it is often the case that there is insufficient expertise to carry out all of the security assessments necessary. The ever increasing number of information systems, which themselves are of increasing complexity and exposed to constantly evolving threats, mean that a greater number of security assessments are required more quickly than ever before. In this research we focus on modelling parts of the assessment process in order to create decision aids that can be used by experts to reduce load, and by non-experts to achieve an approximation of security levels for security components and specific technical attacks.

\subsection{Machine Learning Approaches to Security Assessment}
\label{sec:ML}
Over the years there have been many machine learning approaches to computer security, ranging from classic algorithm based studies \citep{lane2000} to immune system inspired approaches \citep{kim2007}. What almost all of these past approaches have in common is that they attack the problem from a quantitative point of view, based on evaluating datasets such as network traffic , but without an attempt to integrate often highly valuable knowledge held by security experts. Examples include what is probably the earliest use of fuzzy sets in this context \citep{clements1977}, intrusion detection based on multisensor data fusion \cite{bass2000} or fuzzy cognitive maps \cite{siraj2001}, and various automated decision systems to support risk or threat analysis 
\citep{shah2003,ngai2005,linkov2006,sun2006,walle2006,dondo2007,mcgill2007}.
Other approaches include anomaly detection algorithms, e.g. using Bayesian approaches \citep{androutsopoulos2000} to detect spam in emails, or self-organising map algorithms to learn combinations of external signals and system call IDs to detect unusual patterns \citep{feyereisl2012}. Similar to anomaly detection is the area of intrusion detection, which again typically relies on unsupervised machine learning techniques for two class classification approaches of `normal' and `abnormal' behaviour, for example the work by \cite{twycross2010}. 


More general approaches to systematising and supporting methods for the design of secure systems have been reviewed by  \cite{baskerville1993}, \cite{dhillon2001} and more recently by \cite{jansen2009}.
As a refinement of more general approaches, this paper focuses on extracting and making use of and leveraging the insight held by cyber security experts. Experts are commonly good at assessing the security of parts of systems --- such as rating the vulnerability of individual hops or attack paths. However, the major challenges for experts are:
\begin{itemize}
	\item The large number of hops and attack vectors in systems, i.e. the finite number of suitably qualified experts makes the timely assessment of hops and attack paths highly challenging for many users (e.g. companies).
	\item The aggregation of often different ratings for the same components by different experts, e.g. based on different expert backgrounds or levels of expertise.
\end{itemize}

In this context, our paper focuses on alleviating the task load for available experts by introducing, demonstrating and evaluating a novel approach to partially automating the rating of attack vectors based solely on individual hop ratings. The approach proposed employs linear order statistics, specifically the Ordered Weighted Average (OWA), to fuse individual hop ratings into an overall attack path assessments, while the weights of the OWA are determined using an evolutionary algorithm, resulting in overall high quality attack vector vulnerability assessments which closely follow those of experts (if these experts directly assess the vectors). 
The closest body of works in computer security that is related to our approaches is that of attack graphs and scenario creation, e.g. the work by \cite{tedesco2008}. However, in these approaches the aim tends to be to develop improved techniques to construct and rate attack graphs, rather than facilitating and partially automating the overall process of integrating expert knowledge in cyber security assessments. 

\subsection{Weighted and Ordered Weighted Averages}
\label{sec:WOWA}

In the experiments shown in this paper, two aggregation methods are employed: Weighted Averages (WAs), and Ordered Weighted Averages (OWAs) \citep{Yager1988}. OWA operators were chosen for the task of aggregating security component ratings to produce attack rankings following discussions with a group of GCHQ technical experts. In our discussions the hypothesis emerged that the difficulty of a particular attack is largely determined by the most difficult security component to attack or evade; the remaining components contribute to the difficulty in proportion with their own difficulty. Initially, we tried simple maximum operators to see whether they could be used. In practice however, this results in a lot of attacks having tied ranks, providing little insight. This is because, if a particularly difficult component appears in multiple attacks, they will all receive the same rating. OWA operators allow us to assign more weight to the most difficult components, while accounting for the difficulty of the remaining components too. This greatly reduces the potential for tied ranks, and creates meaningful rankings that distinguish between attacks. 

Let $X$ be a set of $N$ information sources (e.g., reviewers, experts, etc.) with each information source contributing evidence $x_i$, $i= {1 \dots N}$.
The standard Weighted Average, WA, combines the information from all the sources by associating the evidence from each source with a given weight. More formally, consider a set $W$ of weights $w_i$ corresponding to each source $x_i$, where $\sum w_i=1$. Then:
\begin{equation}
\me{WA}(X) = \sum_{i=1}^N w_i x_i
\label{eq:WA}
\end{equation}
The Ordered Weighted Average, OWA, combines the information from all the sources, ordered by the size of the evidence (largest to smallest), using a pre-defined vector of weights, $W'$. More formally, consider the vector of evidence $O=(o_1,\dots,o_N)$, formed by the ordering (largest to smallest) of the elements in $X$, such that $o_1 \ge o_2 \ge \dots o_j$. Then:
\begin{equation}
\me{OWA}(X) = \sum_{j=1}^N w'_j o_j
\label{eq:OWA}
\end{equation}
Whereas in the WA, the weights are associated with each source, in the OWA, the weights are associated with an ordered set of the evidence by all sources. Thus, in the OWA, a change in contributed evidence can result in a different ordering and thus in a different mapping of weights to evidence.
For example, in this work we assign a weight to the most difficult component, the second most difficult component and so on. Each of the weights is multiplied by the corresponding component, the first weight is multiplied by the first and thus largest component and so on. If the first weight is near to one, the OWA behaves similarly to a `maximum' operator, thus resulting in the overall assessment being based nearly exclusively on the most difficult component.

OWAs are commonly used in decision making problems to create aggregate scores and/or ratings. In \cite{Canos2008}, OWA operators are used in a personnel selection problem to aggregate selection criteria. In \cite{Badea2011} an OWA is used to compute a rating of the security of energy supply, \cite{Sadiq2010} describes the use of an OWA to aggregate performance indicators of small water utilities creating an overall performance assessment. \cite{Imamverdiev2011} show how a fuzzy OWA operator can be used for rating information security products in terms of reduction of information security risk, and in \cite{Merigo2010} and \cite{Merigo2011} modified OWA operators are applied to a financial product selection problem and a football team player selection problem respectively.

\begin{figure}[!htb]
\centerline{\subfloat[Data Centre and Office Automation]{\label{fig:groupA}\includegraphics[width=\columnwidth]{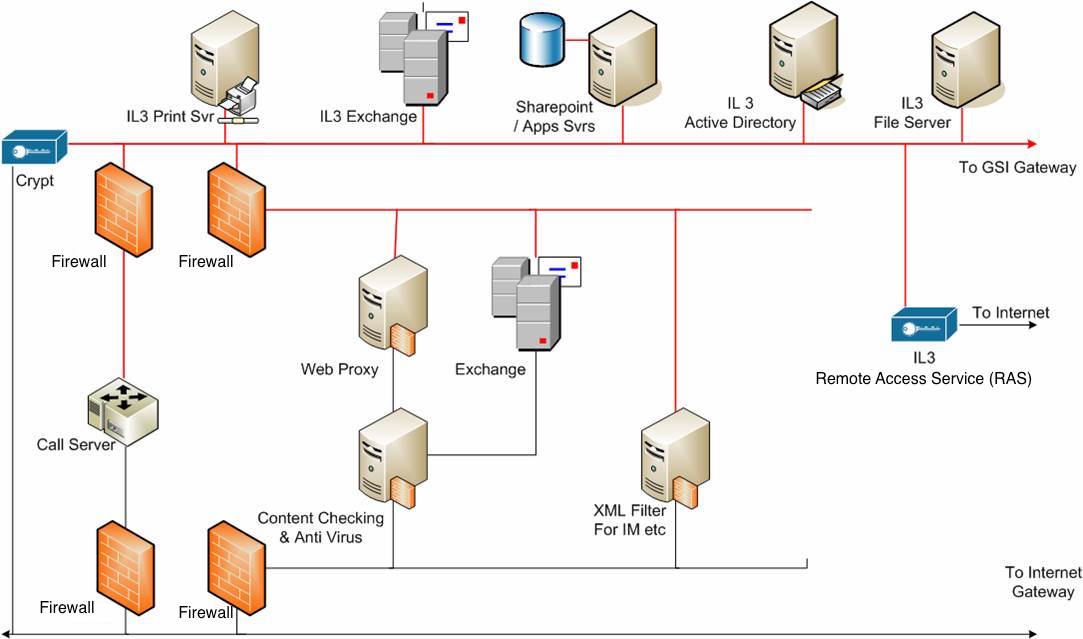}}}
\centerline{\subfloat[Mobile Sites]{\label{fig:groupB}\includegraphics[width=\columnwidth]{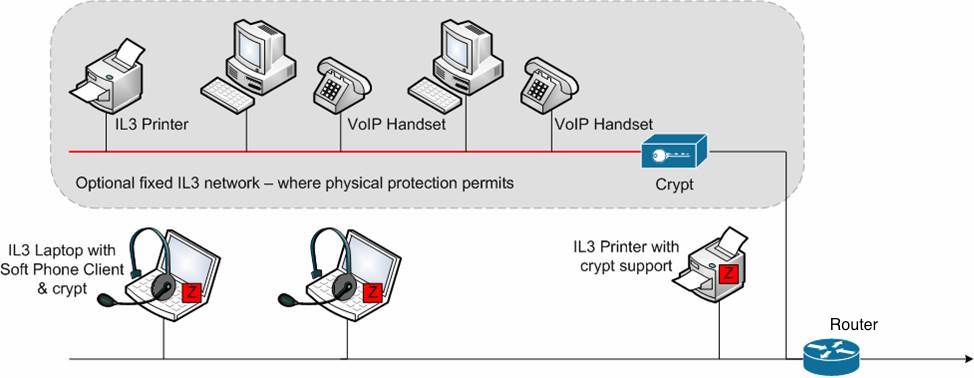}}}
\caption{Scenario System Architecture. Note that symbols are self-explanatory following standard computer network diagram drawing guidelines, based on Cisco icons (see http://www.cisco.com/c/en/us/about/brand-center/network-topology-icons.html).}
\label{fig:Arch}
\end{figure}

\subsection{Evolutionary Algorithms}
\label{sec:EAs}

Evolutionary Algorithms (EAs) are a set of well used heuristic search methods that mimic aspects of biological evolution to \emph{evolve} solutions. The most popular EA, the Genetic Algorithm (GA) (see \cite{Holland1975} and \cite{Goldberg1989}), begins with the generation of a random initial population of solutions to which fitness-based \emph{selection and copy}, \emph{crossover} and \emph{mutation} operators are applied in order to discover near-optimal solutions. The advantage of the GA is that it is able to search a large area of the solution space, while evaluating a small proportion of the possible solutions. This is critical for problems like those seen in this work in which there are far more possible solutions than it is practical to evaluate. The use of a GA to search for suitable OWA weights is demonstrated in \cite{Nettleton2001} and \cite{Torra2011}, and in our own previous work \citep{Miller2013}.

The work in the current paper significantly extends the work seen in \cite{Miller2013} in which EAs were employed to search for appropriate OWA weights to be used for aggregating ratings of security components to produce rankings of technical attacks, and to search for WA weights to be used to aggregate ratings of aspects of security components to produce overall security component ratings. Using an EA in this way not only allows us to discover suitable aggregation operators, it also gives an indication of the relative importance of each component of the aggregation. For example, by looking at the resulting OWA weights, we will be able to confirm or contradict our hypothesis that the most difficult to attack/evade security component contributes the most to the attack ranking. As the EA could theoretically arrive at any OWA, and is led purely by the fitness of an OWA (i.e., how well the resulting rankings match individuals' actual rankings), the OWAs produced are a valid guide to the relative contribution made by each security component. Similarly, the WA weights found will provide insight into which characteristics contribute most to the overall rating, and which have little effect on the overall difficulty of attacking/evading a security component.

\subsection{Statistical Analysis}

To make comparisons between the results produced by the OWA/WA operators and the experts' actual rankings, Spearman's Rho is used. Spearman's Rho \citep{Spearman1904}, also called Spearman's Rank Correlation Coefficent, measures the statistical dependence of two sets of rankings (ordinal data). The coefficient is a number in [-1,1] that indicates the level of correlation; 1 denotes a perfect positive correlation, 0 means there is no correlation, and -1 denotes a perfect negative correlation. To measure the correlation between component characteristics and overall difficulty questions Pearson's $r$ \citep{Pearson1895} is used. Pearson's $r$, also called the Pearson Product-Moment Correlation Coefficient (PPMCC), is a method that measures linear correlation between two sets of values (interval or ratio data); a value of zero indicates no correlation and a value of one indicates perfect correlation. For both methods, generally, values above 0.5 are interpreted as strong correlation.

\subsection{Experimental Validation}

The experimental results and statistical analyses presented in this paper are not tested in a formal manner against, for example, pre-specified null hypotheses or similar. Thus, in statistical terms, it should be viewed as \emph{exploratory data analysis} rather than as formal hypotheses testing (sometimes called \emph{confirmatory data analysis}), as discussed by Tukey in 1962 \citep{tukey1962}. Overall validation of the experimental methods in this paper is presented in terms of the degree to which it is possible to model parts of cyber-security decision makers processes using statistical aggregation of components within attack vectors. As this is exploratory, there are no explicit tests for \emph{falsifiability} --- however, there is an informal null hypothesis that security `experts' are not actually expert, and that no correlations, associations or aggregations could be found that would link rating of sub-components to rankings of attack vectors. But, the main purpose of this work is to carry out exploratory data analysis in this context to explore to what degree it may be possible to submit security assessment to more systematic methods, which in future could lead to new hypotheses with associated methods of new data collection and analysis.

\section{Data Elicitation}
\label{sec:DE}

In this section we will describe the data elicitation exercise conducted at GCHQ, Cheltenham, UK. As the data and actual elicitation is vital to the design and functionality of the automated reasoning system introduced later, we proceed by describing the elicitation process and resulting data in detail. Thirty nine highly experienced GCHQ selected cyber security experts from government and commercial backgrounds including system and software architects, technical security consultants, penetration testers, vulnerability researchers and specialist systems evaluators took part in a decision making exercise. While the detail of the actual attack scenarios and security components is not available for general publication, we provide the actual numeric `anonymised' data in the paper.

\subsection{Scenario}

For the purposes of the exercise, a scenario was created and presented by senior technical staff at GCHQ. The scenario was designed to be a typical example of a government system. The system has a range of core services and back end office facilities, along with remote sites and mobile access. Figure \ref{fig:Arch} provides an overview of the system architecture. The most sensitive information is held in core systems, with assets rated at Business Impact Level 3 (BIL3). Business Impact Levels are the UK Government's standard classification scheme for rating the ramifications of assets being disseminated, altered or destroyed, ranging from no consequences (BIL0) to catastrophic consequences (BIL6).

The experts were presented with information regarding the architecture of the system and its components, and allowed to ask questions. In real-world assessments, a great deal of information about a system and its components is required in order to perform a security assessment, including details of hardware and software, version/revision numbers and the frequency with which updates and patches are applied. In order to address this in the scenario, without overcomplicating the exercise (making it infeasible), the experts were asked to make assumptions about the system, specifically, that the software/hardware used was typical of this type of government system and that security policy was being applied in manner consistent with how they typically find it in their day-to-day work. This was acceptable for the experts, as they all have considerable experience with the type of BIL3 government system presented in the scenario, and how security policies are generally applied.

\begin{figure}[!htb]
\centerline{\subfloat[Data Centre and Office Automation]{\label{fig:AV_DC}\includegraphics[width=\columnwidth]{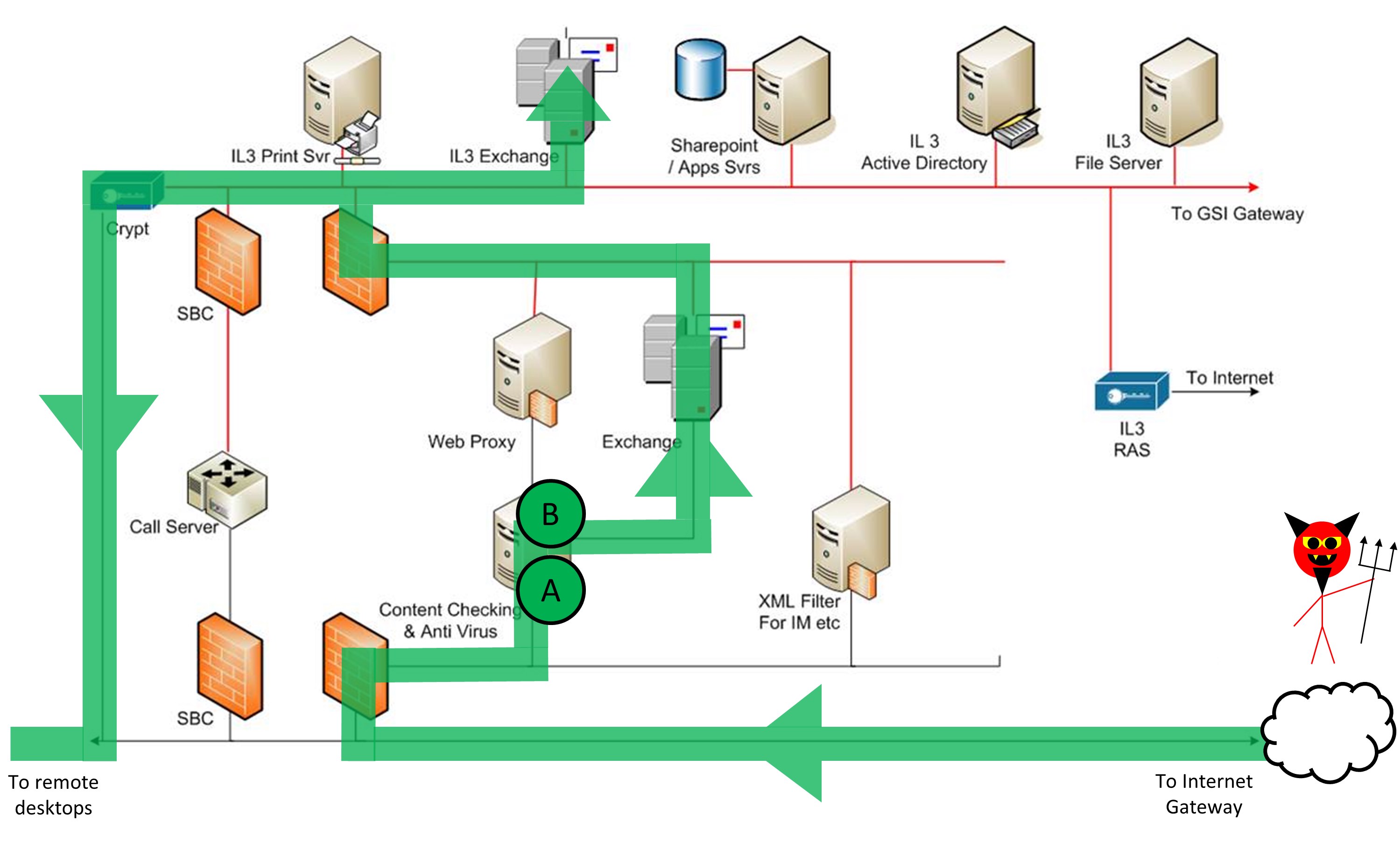}}}
\centerline{\subfloat[Mobile Sites]{\label{fig:AV_MS}\includegraphics[width=\columnwidth]{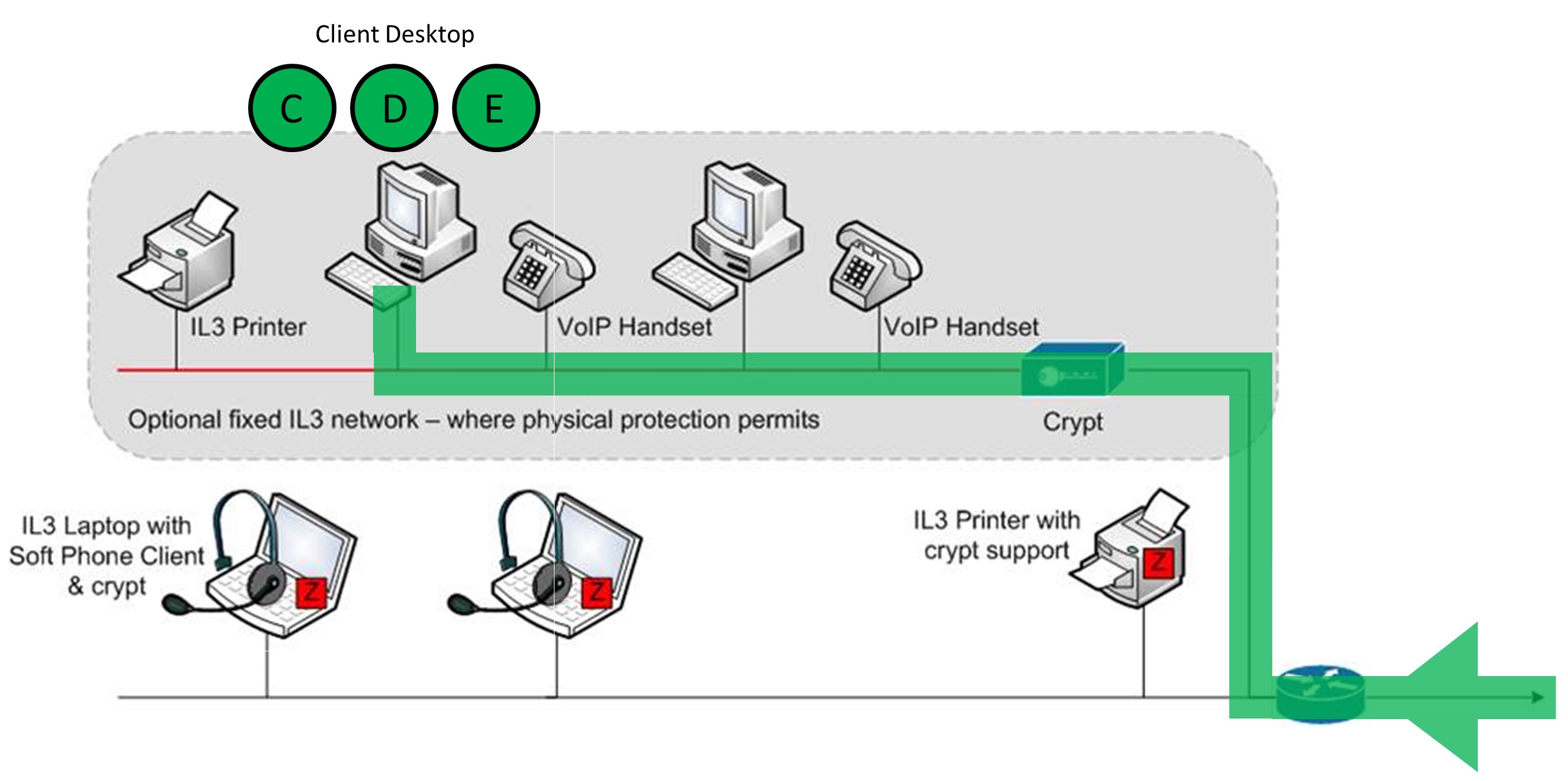}}}
\caption{Example Technical Attack from Scenario}
\label{fig:AV}
\end{figure}

In addition to details of the system, the experts were also given a list of 10 technical attacks on the system, and the 26 security components that make up the attacks. Figure \ref{fig:AV} shows an example of the information the experts were given about the technical attacks. In this attack an email with a malicious attachment is sent to a recipient within the system, when opened the attachment runs an exploit taking control of the recipient's PC, allowing the attacker to launch further attacks from their machine. Figure \ref{fig:AV_DC} shows the path the attacker takes from outside the system, through the back office systems. Once this is achieved the attacker proceeds to the mobile sites as shown in Fig. \ref{fig:AV_MS}, completing the attack by compromising a client desktop.

\begin{table}[!htb]
\begin{center}
	\caption{Components in Example Attack}
	\label{tab:Hops}
	\begin{tabular}{c|c c}
		\hline
		\textbf{Step} & \textbf{Attack/Evade} & \textbf{Component}\\
		\hline
		A & Evade & Content Checker\\
		B & Evade & Anti-Virus Software\\
		C & Attack & PDF Renderer\\
		D & Evade & Anti-Virus Software\\
		E & Attack & Client Access Controls\\				
		\hline
	\end{tabular}
\end{center}
\end{table}

The five steps in the attack are labelled A to E in Fig. \ref{fig:AV}, Table \ref{tab:Hops} provides details of the security components involved. Each of these components must be attacked (compromised) or evaded (bypassed) in order for the attack to be successful. After being shown the 10 attacks and their constituent components, the experts were asked to complete two survey exercises:

\begin{enumerate}
	\item Rank the 10 technical attacks in order of difficulty to complete without being noticed.
	\item Rate the 26 security components in terms of:
	\begin{enumerate}
		\item The overall difficulty of attacking or evading the component.
		\item Individual factors that contribute to the difficulty of attacking or evading the component.
	\end{enumerate}
\end{enumerate}

Each activity was conducted in silence, with experts sat at separate desks, to avoid experts influencing each others' opinions. Ranking the technical attacks like this is an intuitive way of establishing the relative difficulty of completing each of the attacks successfully. Experts make direct comparisons between attacks. e.g., \emph{`Is attack $x$ more or less difficult than attack $y$?'}, in order to arrive at a ranked list. This approach can make it easier for experts to identify salient attacks, as in some cases it is difficult to place a precise value on the difficulty of an attack.

When rating security components, experts were asked to give a rating on a scale from 0 to 100 to the question \emph{`Overall, how difficult would it be for an attacker to successfully attack/evade this component without being noticed?'}, and a series of questions about factors that contribute to the overall difficulty of attacking/evading a component. These questions were created by GCHQ technical experts, who identified what they believed to be the important factors that contribute to the difficulty of attacking/evading security components. The rating scale from 0 to 100 was completely arbitrary, with no absolute meaning of 0 or 100 being conveyed to the experts, and the response included both a perceived location (`mean') and uncertainty (`variance') of opinion as described in more detail in \cite{Miller2013a}.

Two sets of questions were used, one for when an attack required a component to be attacked (compromised), and another for when an attack required a component only to be evaded (bypassed). The categorisation of components into `attack' or `evade' was made by the GCHQ technical team who designed the scenario. While it is possible that some components may in practice be either attacked or evaded, and that different attach vectors may be possible, such alternatives were not considered. That is, the list of attack vectors created, and whether the components were compromised or bypassed within such an attack, were designed and specified by the GCHQ technical team.

 The questions asked when a component needed to be attacked were:
\begin{itemize}
	\item How complex is the target component (e.g., in terms of size of code, number of sub-components)?
	\item How much does the target component process/interact with any data input?
	\item How often would you say this type of attack is reported in the public domain?
	\item How likely is it that there will be a publicly available tool that could help with this attack?
	\item How inherently difficult is this type of attack? (i.e., how technically demanding would it be to do from scratch, with no toolset to help.)
	\item How mature is this type of technology?
	\item How easy is it to carry this attack out without being noticed?
	\item Overall, how difficult would it be for an attacker to do this?
\end{itemize}

For components that needed only to be evaded, the following questions were asked:
\begin{itemize}
	\item How complex is the job of providing this kind of defence?
	\item How likely is it that there will be publicly available information that could help with evading this defence?
	\item How mature is this type of technology? 
	\item Overall, how difficult would it be for an attacker to do this?
\end{itemize}

This type of questioning allows experts to give an indication of the absolute difference in difficulty between security components, unlike ranking which provides an intuitive way to obtain an ordering but elicits no information about the degree of difference between attacks. 

The dataset created by this exercise is made up of three levels:
\begin{enumerate}
	\item Rankings of attacks.
	\item Ratings of the overall difficulty of attacking/evading specific security components that make up attacks.
	\item Ratings of characteristics/factors that contribute to the overall difficulty of attacking/evading specific security components.
\end{enumerate}

Details of analysis work carried out previously on the resulting data can be found in \cite{Miller2013a}. 

%

The exercise and methods described in this paper represent an important part of performing a security assessment on a proposed system. The process of rating security components and highlighting the easiest ways to attack a system via its weakest components is critical to understanding the levels of risk posed by a system design. In this study we will show how rankings of attacks can be produced from overall ratings of components, how overall ratings of components can be created from ratings of characteristics of components, and how the two stages can be chained together to produce attack rankings from ratings of characteristics of components.

\section{Data Analysis Methods}
\label{sec:imp}


\subsection{OWA/WA Aggregation}

The methodology in this paper is based on determining the `optimum' weights for either WA or OWA aggregation of the assessment of difficulty of compromising the various components in a range of attack vectors. This in turn, enables both the ranking of attack vectors in terms of their overall difficulty as well as providing an understanding of exactly which components in a security system contribute the most/least to an overall system being compromised. Below, we briefly summarise the methodology and detailed rationale for both the WA and the OWA.

In all, 26 security components were rated, each belongs to one or more of the 10 specified technical attacks. Table 2 lists the attacks, and their constituent security components. Notice that some components appear more than once in the same attack, this is because in some cases more than one instance of a component needs to be attacked or evaded for an attack to be completed successfully. For example, an attacker may have to compromise multiple firewalls in order to obtain access to their target. The order of components is important --- alternative orderings, with perhaps alternative attack methodologies, may be possible in the real world, but these would be considered as different overall attacks.

\begin{table}[!htb]
\begin{center}
	\caption{Attacks and their Constituent Security Components}
	\label{tab:AVHops}
	\begin{tabular}{c|c}
		\hline
		\textbf{Attack} & \textbf{Components}\\
		\hline
		1 & 2,3,1,4,5\\
		2 & 6,7,6,8,4\\
		3 & 9 \\
		4 & 10,11,4,5\\
		5 & 12,13,2,3,14,15,4,5\\
		6 & 16,16,17,4,5\\
		7 & 6,18,4,5\\
		8 & 19,20,21\\
		9 & 22,23,24\\
		10 & 25,26,1,4,5\\
		\hline
	\end{tabular}
\end{center}
\end{table}

With the information regarding the composition of attacks, a WA can be employed for the aggregation of the ratings  of the difficulty of attacking / evading a security component, to produce an overall difficulty rating. In the instance of the WA, weights are applied to specific factors, e.g., the complexity of a component. The result is a set of derived overall difficulty ratings for all 26 security components. The core challenge here is the search for the `optimal' weights of the given WA. To address this, this paper proposes the use of Evolutionary Algorithms (EAs) which will be discussed in the next section.

Similarly, an OWA is used to produce attack rankings from overall difficulty assessments of their constituent hops. However, as noted in Section~\ref{sec:WOWA}, the OWA matches a vector of weights to the ordered set of evidence (difficulties) for the individual components. For instance, an OWA can be used to aggregate overall difficulty ratings of components to produce a rating for each of the ten attacks. As an example, Table 3 shows the overall difficulty ratings that an expert gave for attacking/evading the security components in Attack 1. 

If we want to aggregate these values into an overall rating for Attack 1 that gives more weight to the most difficult components, we could choose the weights \[ W' =  ( 0.33, 0.27, 0.20, 0.13, 0.07 ). \] The next step is to order the ratings \[ X = \{ 25.00, 40.00, 20.50, 40.00, 70.00 \} \] by difficulty to create the ordering \[ O = ( 70.00, 40.00, 40.00, 25.00, 20.50 ). \] Equation~\ref{eq:OWA} shows how the OWA can then be applied by multiplying each weight by its corresponding element in the ordering. In this case the resulting value is 46.59. When we compare this with the mean value 39.10, it can be seen that this OWA operator has given more weight to those components with higher difficulty ratings. If we then repeated this process with the remaining nine attacks, a complete set of attack ratings would be produced that could be used to generate attack rankings.

\begin{table}[!htb]
\begin{center}
	\caption{Overall Difficulty Ratings for the Security Components in Attack 1}
	\label{tab:Attack1}
	\begin{tabular}{c|c}
		\hline
		\textbf{Component} & \textbf{Overall Difficulty}\\
		\hline
		1 & 25.00\\
		2 & 40.00\\
		3 & 20.50\\
		4 & 40.00\\
		5 & 70.00\\
		\hline
	\end{tabular}
\end{center}
\end{table}

\begin{table*}[!htb]
\begin{center}
	\caption{EA-OWA: EA Configuration - Best OWAs}
	{\begin{tabular}{c|c c c c c c c c}
		\hline
		& \multicolumn{8}{c}{\textbf{Best Weights}}\\
		\textbf{Test} & \textbf{1}& \textbf{2}& \textbf{3}& \textbf{4}& \textbf{5}& \textbf{6}& \textbf{7}& \textbf{8}\\
		\hline
		\multicolumn{9}{c}{\textbf{Even}}\\
		\hline	
		12 & 0.8907 & 0.0011 & 0.0323 & 0.0120 & 0.0070 & 0.0174 & 0.0035 & 0.0359\\
		13 & 0.8899 & 0.0044 & 0.0297 & 0.0126 & 0.0078 & 0.0242 & 0.0030 & 0.0284\\		
		\hline
		\multicolumn{9}{c}{\textbf{Odd}}\\
		\hline
		13 & 0.7858 & 0.0223 & 0.1285 & 0.0471 & 0.0115 & 0.0031 & 0.0015 & 0.0002\\
		16 & 0.7854 & 0.0246 & 0.1248 & 0.0479 & 0.0117 & 0.0033 & 0.0001 & 0.0021\\
		\hline
		\multicolumn{9}{c}{\textbf{All}}\\
		\hline
		1 & 0.9582 & 0.0028 & 0.0242 & 0.0003 & 0.0057 & 0.0029 & 0.0002 & 0.0056\\
		2 & 0.9621 & 0.0017 & 0.0221 & 0.0003 & 0.0045 & 0.0059 & 0.0009 & 0.0025\\
		\hline
	\end{tabular}}
	\label{tab:EAConfigOWA}
\end{center}
\end{table*}

\subsection{EA Implementation}

In this work, EAs (as introduced in Section \ref{sec:EAs}) are used to optimise the weights for:
\begin{enumerate}
\item WAs, used to aggregate ratings of difficulty of factors to produce the overall difficulty ratings of security components;
\item OWAs, used to aggregate the overall difficulty ratings of security components to create rankings of the overall attack vectors.
\end{enumerate}
The EA has been implemented for both the WA and OWA cases as follows.

\emph{Solutions} are represented as a vector of weights. In this example, the maximum number of security components in an attack is eight, therefore each vector contains eight weights. Those attacks that do not have eight components are padded out with zeros. This is to avoid concentrating weightings (which must add up to 1) on fewer components, which can have the effect of making an attack more difficult purely because the weights are concentrated on fewer component ratings. In general, an attack should get more difficult as components are added to it.

An \emph{initial population} of individuals is created at the beginning of the algorithm. Each individual is produced by generating seven random points on a line from 0 to 1. From this, eight weights are created by taking the distances between 0, each of the seven points, and 1. This ensures that the result is eight values that add up to 1.

To compute \emph{fitness} each expert's actual attack ranking is compared with the attack ranking derived from the overall secu-rity component rating with the current OWA. The comparison is made using Spearman's rank correlation coefficient, which produces a value between -1 (perfect negative correlation), 0 (no correlation) and 1 (perfect positive correlation). An error value is then calculated by subtracting the correlation coefficient from 1, so a perfect positive correlation produces an error of 0, and anything less produces an error value greater than 0. Once an error value has been calculated for each expert, the Mean Squared Error (MSE) is computed to give an error value for each solution in the population.

\emph{Selection} is achieved by sorting the population of solutions by fitness, and then generating the indexes of selected parents from a complementary cumulative distribution. A lower numbered individual is more likely to be chosen, ensuring that fitter individuals a more likely to be selected, though potentially any individual can be selected.

To guarantee that the fittest individual in successive generations is not worse that in preceding generations \emph{elitism} selects the best individuals from the current generation and copies them unaltered into the next generation. A similar operator is the the \emph{copy} operator, which takes a parent picked using the described selection method and copies them into the next generation.

The \emph{mutation} operator used in this EA implementation randomly selects two weights from a solution, increases one by a small amount and reduces the other by the same amount. This ensures that the weights still add up to one afterwards. The resulting weights are validated, if either becomes greater than 1 or less than 0, another two elements are selected.

The \emph{crossover} method used is a single point crossover. Two parents are selected and a child is created that consists of the first four weights of the first parent, and the last four from the second parent. It is unlikely at this point that the weights will add up to 1, so they are normalised. This method of crossover ensures valid OWAs, while preserving the characteristics of each contributing parent.

Finally, the EA's \emph{termination criteria} is set to be a specific number of generations, after which the algorithm stops.

%
%

\section{From Factor Ratings to Security Component Ratings}
\label{sec:Comp}

In Section~\ref{sec:imp}, we described how an EA has been implemented for use in this research. In this section we employ the described EA to perform a search for WA weights. In these experiments, we show how overall difficulty ratings for security components can be obtained from ratings of their characteristics using WAs where the weights have been discovered using an EA. As well as producing overall difficulty ratings, this process will also highlight which questions contribute most/least to the overall ratings. In this case it is not rankings that are being compared, it is ratings. Because of this, Spearman's rho is not  an appropriate method for comparison so an alternative method of calculating fitness was established. The `error' is directly calculated by taking the sum of absolute differences between the derived hop ratings and the experts' actual hop ratings. Each expert's error value is then used to calculate a MSE value for all experts, which is used as the fitness value for each solution.

When performing data analysis with the ratings of characteristics/factors we found that the relationship between them and the overall difficulty of attacking/evading a security component varies between question sets. This is because they contain different numbers of questions, and some questions are specific to a question set. Because of this, we will perform separate experiments with the questions for attacking and evading security components, finding WAs specific to each.

\subsection{EA Configuration}

Initially, a series of tests were conducted to find ideal configurations of the EA. Full details of these experiments and their results can be found in \ref{sec:AppA_WA}. The best results for each question set can be found in Table \ref{tab:WAConfigRes}; the best EA configurations are shown in Table \ref{tab:WAConfigEA}. The best EA found for each question set is used in the next subsection for extended experiments.

\begin{table}[!htb]
\begin{center}
	\caption{EA-WA: EA Configuration - Best Results}
	{\begin{tabular}{c|c c}
		\hline
		& \multicolumn{2}{c}{}\\
		\textbf{Test} & \textbf{Mean Sp.} & \textbf{MSE}\\
		\hline
		\multicolumn{3}{c}{\textbf{Attack}}\\
		\hline	
		15 & 0.7931 & 194.0125\\	
		\hline
		\multicolumn{3}{c}{\textbf{Evade}}\\
		\hline
		1 & 0.2270 & 420.2156\\
		\hline
	\end{tabular}}
	\label{tab:WAConfigRes}
\end{center}
\end{table}

\begin{table}[!htb]
\begin{center}
	\caption{EA-WA: EA Configuration - Best EAs}
	{\begin{tabular}{c|c|c|c|c|c}
		\hline
		\multicolumn{6}{c}{\textbf{Attack}}\\
		\hline	
		\textbf{Test} & \textbf{Gens} & \textbf{Pop} & \textbf{Copy} & \textbf{Cross} & \textbf{Mut}\\
		\hline
		15 & 400 & 155 & 0.20 & 0.20 & 0.59\\
		\hline
		\multicolumn{6}{c}{\textbf{Evade}}\\
		\hline	
		\textbf{Test} & \textbf{Gens} & \textbf{Pop} & \textbf{Copy} & \textbf{Cross} & \textbf{Mut}\\
		\hline
		1 & 250 & 250 & 0.00 & 0.20 & 0.79\\
		\hline
	\end{tabular}}
	\label{tab:WAConfigEA}
\end{center}
\end{table}

\subsection{Extended Experiments}

In these extended experiments the best EA configuration found for each question set will be used in 30 runs with differing random seeds. The purpose of these experiments is to assess the consistency of the EAs, and to see the best results we can reasonable expect with the EA. In experiments with the attack question set the best configuration (Test 15) will be used; \ref{sec:AppA_WA} shows that multiple configurations achieved the best result for the evade question set, of these we have arbitrarily chosen to use the configuration from Test 1. Table \ref{tab:EAWA30runsComp} shows a summary of the results for the attack question set. The results are extremely stable, there is little variation in the results over the 30 runs. Table \ref{tab:EAWA30RunsWAComp} gives details of the best WA found.

\begin{table}[!htb]
\begin{center}
	\caption{EA-WA: Extended Experiments - Attack}
	{\begin{tabular}{c|c|c|c}
		\hline
		\multicolumn{4}{c}{\textbf{Spearman's Rho}}\\
		\hline
		\textbf{Max} & \textbf{Min} & \textbf{Mean} & \textbf{Standard Deviation}\\
		\hline
		0.7940 & 0.7919 & 0.7935 & 0.0005\\
		\hline
		\multicolumn{4}{c}{\textbf{MSE}}\\
		\hline
		\textbf{Max} & \textbf{Min}& \textbf{Mean} & \textbf{Standard Deviation}\\
		\hline
		194.2998 & 193.9896 & 194.0856 & 0.0741\\
		\hline
	\end{tabular}}
	\label{tab:EAWA30runsComp}	
\end{center}
\end{table}

\begin{table}[!htb]
\begin{center}
	\caption{EA-WA: Extended Experiments - Best WA - Attack}
	{\begin{tabular}{c c c c}
		\hline
		\multicolumn{4}{c}{\textbf{Best Weights}}\\
		\textbf{Test} & \textbf{Complexity}& \textbf{Interaction}& \textbf{Frequency}\\
		\hline
		22 & 8.43E-05 & 0.0012 & 0.1622\\		
		\hline
		\textbf{Tool}& \textbf{Inherent}& \textbf{Maturity}& \textbf{Unnoticed}\\
		\hline
		0.2189 & 0.4020 & 0.0438 & 0.1719\\		
		\hline
	\end{tabular}}  
	\label{tab:EAWA30RunsWAComp}
\end{center}
\end{table}

\begin{table*}[!htb]
\begin{center}
	\caption{EA-WA: Extended Experiments - Question Ranking - Attack}
	{\begin{tabular}{c|c|c|c}
		\hline
		\textbf{Rank} & \textbf{Question} & \textbf{Weight} & \textbf{Mean Corr}\\
		\hline
		1 & How inherently difficult is this type of attack? & 0.4020 & 0.6454\\
		2 & How likely is it that there will be a publicly available tool that could help with this attack? & 0.2189 & 0.6732\\
		3 & How easy is it to carry this attack out without being noticed? & 0.1719 & 0.4543\\
		4 & How often would you say this type of attack is reported in the public domain? & 0.1622 & -0.6268\\
		5 & How mature is this type of technology? & 0.0438 & 0.1701\\
		6 & How much does the target component process/interact with any data input? & 0.0012 & -0.2741\\
		7 & How complex is the target component? & 8.43E-05 & -0.0623\\
		\hline
	\end{tabular}}
	\label{tab:EAWAQOrder}
\end{center}
\end{table*}

The results of the experiments with the attack question set show that there are clear differences in the relative importance of each question in deriving the overall difficulty of a security component. In order, the questions are ranked as shown in Table \ref{tab:EAWAQOrder}. For comparison, the mean (over all 39 experts) Pearson's r for the correlation between the ratings given for each characteristic, and the overall difficulty is also given. The weightings shown are taken from the test that resulted in the lowest MSE (Test 22), though all tests produced similar results.

The \emph{inherent difficulty} rating has the highest weighting, and it is also strongly correlated with the experts' overall ratings for security components. This may be because the experts do not make a strong distinction between the inherent difficulty of attacking a security component and the overall difficulty, as the terms are similar. The weights attributed to the ratings regarding factors of component difficulty are in line with the correlation between experts' ratings of each factor and their overall rating for each component. In view of the weightings and correlations, it is reasonable to say that the questions ranked 1 to 4 have a demonstrable relationship with how experts rate the overall difficulty of compromising a security component. There does not appear to be a significant relationship between how experts rated the factors in questions 5 to 7 and how they rated the overall difficulty of compromising security components.

From this we could infer that these three questions about the security components are not useful when trying to determine the difficulty of compromising a component, and that only the four with high correlation/weighting are necessary. Alternatively, it could be that the four high correlation/weighting questions were the only ones that experts are able interpret and answer in a consistent manner with the information given. Regardless, using the four questions overall ratings that are closely correlated with participants'€™ actual ratings of overall difficulty can be produced.

Table \ref{tab:EAWA30RunsBy} provides a summary of the results with the evade question set. 
Like the experiments with the attack question set, the results are extremely stable. Over 30 runs, 28 produced the same MSE. Table \ref{tab:EAWA30RunsWABy} shows the WA found in Test 1, though all weightings are within 0.0003 of those shown.

\begin{table}[!htb]
\begin{center}
	\caption{EA-WA: Extended Experiments - Evade}
	{\begin{tabular}{c|c|c|c}
		\hline
		\multicolumn{4}{c}{\textbf{Spearman's Rho}}\\
		\hline
		\textbf{Max} & \textbf{Min} & \textbf{Mean} & \textbf{Std Dev}\\
		\hline
		0.2273 & 0.2270 & 0.2270 & 5.38E-05\\
		\hline
		\multicolumn{4}{c}{\textbf{MSE}}\\
		\hline
		\textbf{Max} & \textbf{Min}& \textbf{Mean} & \textbf{Std Dev}\\
		\hline
		420.2158 & 420.2156 & 420.2156 & 4.76E-05\\
		\hline
	\end{tabular}}
	\label{tab:EAWA30RunsBy}
\end{center}
\end{table}

\begin{table}[!htb]
\begin{center}
	\caption{EA-WA: Extended Experiments - Best WA - Evade}
	{\begin{tabular}{c|c c c}
		\hline
		& \multicolumn{3}{c}{\textbf{Best Weights}}\\
		\textbf{Test} & \textbf{Complexity}& \textbf{Information}& \textbf{Maturity}\\
		\hline
		1 & 0.2185 & 0.4971 & 0.2845\\		
		\hline
	\end{tabular}}
	\label{tab:EAWA30RunsWABy}
\end{center}
\end{table}

Again, some questions are consistently given more weight than others. Table \ref{tab:EAWAQOrderBy} provides a list of the difficulty factor questions ranked by the weight they were given by the EA. The Mean Corr column in Table \ref{tab:EAWAQOrderBy} gives the mean Pearson's r value for the correlation between experts' ratings of factors and their overall ratings. When compared with the correlation values for the attack questions, these are much lower, which may explain why the EA is unable to find a WA that produces overall ratings that are strongly correlated with experts' actual overall ratings.

\begin{table*}[!htb]
\begin{center}
	\caption{EA-WA: Extended Experiments - Question Ranking - Evade}
	{\begin{tabular}{c|c|c|c}
		\hline
		\textbf{Rank} & \textbf{Question} & \textbf{Weight} & \textbf{Mean Corr}\\
		\hline
		1 & How likely is it that there will be publicly available & 0.4971 & -0.3307\\
		 & information that could help with evading defence? &  & \\
		2 & How mature is this type of technology? & 0.2845 & -0.0348\\
		3 & How complex is the job of providing this kind of defence? & 0.2185 & 0.0278\\
		\hline
	\end{tabular}}
	\label{tab:EAWAQOrderBy}
\end{center}
\end{table*}

\section{From Security Component Ratings to Attack Rankings}
\label{sec:Attack}

In this section, we describe how the EA was employed to perform a search for OWA weights. The OWAs are to be used for aggregating overall difficulty ratings of security components to produce difficulty rankings of attacks. In these experiments three groupings of experts have been used: \emph{odd} experts i.e., experts assigned an odd number, \emph{even} experts i.e., experts assigned an even number and \emph{all} experts. This has been done so that OWAs discovered with one group (e.g., \emph{odd}) can be applied to the alternate group (e.g., \emph{even}) to assess their robustness, and the \emph{all} group can be used to obtain the best OWAs that can be found for the entire group of experts for comparison purposes. Extended experiments are used with the best configuration found with the \emph{all} group to test the stability of the configuration, and to further explore the best OWAs that can be found.

\subsection{EA Configuration}

Initially, a series of experiments was conducted to find ideal configurations of the EA for use in these tests. Full details of these experiments and their results can be found in \ref{sec:AppA_OWA}. The best OWAs found are shown in Table \ref{tab:EAConfigOWA} and the best results from these experiments are shown in Table \ref{tab:EAConfigRes}. The best OWAs found for the \emph{odd} and \emph{even} groups will be assessed for their robustness, and the best EA configuration for the \emph{all} group will be used in extended experiments.

\begin{table}[!htb]
\begin{center}
	\caption{EA-OWA: EA Configuration - Best Results}
	{\begin{tabular}{c|c c}
		\hline
		\textbf{Test} & \textbf{Mean Sp.} & \textbf{MSE}\\
		\hline
		\multicolumn{3}{c}{\textbf{Even}}\\
		\hline	
		12 & 0.6885 & 0.1421\\
		13 & 0.6866 & 0.1421\\		
		\hline
		\multicolumn{3}{c}{\textbf{Odd}}\\
		\hline
		13 & 0.5732 & 0.2519\\
		16 & 0.5733 & 0.2519\\
		\hline
		\multicolumn{3}{c}{\textbf{All}}\\
		\hline
		1 & 0.6159 & 0.2069\\
		2 & 0.6159 & 0.2069\\
		\hline
	\end{tabular}}
	\label{tab:EAConfigRes}
\end{center}
\end{table}

\subsection{OWA Robustness}

In order to assess the robustness of the OWAs found by the EA, the best OWAs found for the \emph{odd} group were applied to the \emph{even} group and vice versa. The results of these experiments are provided in Table \ref{tab:EAoddeven}.  It can be seen that while there is some degradation of performance when the OWAs are applied to the alternate group, they produce rankings that are strongly correlated with experts' actual rankings (i.e., Spearman's Rho $>$ 0.5 indicating strong correlation) suggesting that they are robust. As these OWAs work well on their alternate (unseen) group, it is reasonable to expect that OWAs trained with sample data will work well on unseen data.

\begin{table}[!htb]
\begin{center}
	\caption{EA-OWA: OWA Robustness Experiments}
	{\begin{tabular}{c|c c}
		\hline
		& \multicolumn{2}{c}{} \\
		\textbf{Test} & \textbf{Mean Sp.} & \textbf{MSE}\\
		\hline
		\multicolumn{3}{c}{\textbf{Best \emph{even} OWA applied to \emph{odd}}}\\
		\hline
		12 & 0.5228 & 0.3013\\
		13 & 0.5146 & 0.3118\\
		\hline
		\multicolumn{3}{c}{\textbf{Best \emph{odd} OWA applied to \emph{even}}}\\
		\hline
		13 & 0.5915 & 0.2381\\
		16 & 0.5947 & 0.2337\\
		\hline
	\end{tabular}}
	\label{tab:EAoddeven}
\end{center}
\end{table}

\vspace{-0.8cm}

\subsection{Extended Experiments}

The final set of experiments take the best configuration found with the \emph{all} grouping and conduct 30 runs with varying random seeds to allow assessment of the stability of the EAs, and to give a better picture of the best OWAs that can reasonably be expected using the proposed approach. The EA configuration used is shown in Table \ref{tab:EAConfig}. A summary of the results of the extended experiments can be found in Table \ref{tab:EA30runs}. The summary shows that with 30 different random seeds, the results are very stable, there is minimal difference between the solutions found and they all result in OWAs that produce rankings that are strongly correlated with experts' actual rankings. Also of interest is that in all 30 runs the solutions found never place less than 0.92 of the weight on the most difficult to attack/evade security component, corroborating our hypothesis that this is the most important component when assessing the difficulty of an attack.

\begin{table}[!htb]
\begin{center}
	\caption{EA-OWA: Extended Experiments - EA Configuration}
	{\begin{tabular}{c|c|c|c|c}
		\hline
		\textbf{Gens} & \textbf{Pop} & \textbf{Copy}& \textbf{Cross}& \textbf{Mut}\\
		\hline
		250 & 250 & 0.00 & 0.20 & 0.79\\
		\hline
	\end{tabular}}
	\label{tab:EAConfig}
\end{center}
\end{table}

\begin{table}[!htb]
\begin{center}
	\caption{EA-OWA: Extended Experiments}
	{\begin{tabular}{c|c|c|c}
		\hline
		\multicolumn{4}{c}{\textbf{Spearman's Rho}}\\
		\hline
		\textbf{Max} & \textbf{Min} & \textbf{Mean} & \textbf{Std Dev}\\
		\hline
		0.6200 & 0.6104 & 0.6150 & 0.0022\\
		\hline
		\multicolumn{4}{c}{\textbf{MSE}}\\
		\hline
		\textbf{Max} & \textbf{Min}& \textbf{Mean} & \textbf{Std Dev}\\
		\hline
		0.2137 & 0.2057 & 0.2092 & 0.0020\\
		\hline
	\end{tabular}}
	\label{tab:EA30runs}
\end{center}
\end{table}

\section{From Factor Ratings to Attack Rankings}
\label{sec:All3}

In this section we will use the best WAs and OWAs, identified as described in the previous sections, to derive attack rankings from the difficulty factor ratings. This is a two stage process: firstly, the best WAs will be used to compute overall difficulty ratings for security components from ratings of factors of that difficulty; secondly, the derived overall difficulty ratings will be aggregated using the best OWA to create attack rankings. Table \ref{tab:BestWAOWA} shows the WA and OWA operators used, and Table \ref{tab:WAOWAResults} gives a summary of the results. Note that in this case pairs of rankings are being compared, as such, error is calculated by subtracting the Spearman's Rho value from 1, as in Section \ref{sec:Attack}. Because of this it can be seen that, for example, the minimum error is the complement of the maximum Spearman's Rho as they refer to the same individual. In the previous experiments where overall difficulty ratings for security components were being compared, error was calculated by taking the absolute value of the differences between the actual ratings provided by the experts, and the ratings derived from the difficulty factor ratings.

\begin{table}[!htb]
\begin{center}
	\caption{Factor Ratings to Attack Rankings: WAs and OWA}
	{\begin{tabular}{c|c c|c}
		\hline
		& \multicolumn{2}{c|}{\textbf{WA}} & \\
		\textbf{Index} & \textbf{Attack} & \textbf{Evade} & \textbf{OWA}\\
		\hline
		1 & 8.43E-05 & 0.2185 & 0.9484\\
		2 & 0.0012 & 0.4971 & 6.7E-05\\
		3 & 0.1622 & 0.2845 & 0.0368\\
		4 & 0.2189 & - & 0.0120 \\
		5 & 0.4020 & - & 0.0008\\
		6 & 0.0438 & - & 0.0016\\
		7 & 0.1719 & - & 0.0002\\
		8 & - & - & 0.0002\\
		\hline
	\end{tabular}}
	\label{tab:BestWAOWA}
\end{center}
\end{table}


\begin{table}[!htb]
\begin{center}
	\caption{Factor Ratings to Attack Rankings: Results}
	{\begin{tabular}{c|c|c|c}
		\hline
		\multicolumn{4}{c}{\textbf{Spearman's Rho}}\\
		\hline
		\textbf{Max} & \textbf{Min} & \textbf{Mean} & \textbf{Std Dev}\\
		\hline
		0.9636 & -0.4303 & 0.5562 & 0.3387\\
		\hline
		\multicolumn{4}{c}{\textbf{Error}}\\
		\hline
		\textbf{Max} & \textbf{Min}& \textbf{MSE} & \textbf{Std Dev}\\
		\hline
		1.4303 & 0.0364 & 0.3117 & 0.3387\\
		\hline
	\end{tabular}}
	\label{tab:WAOWAResults}
\end{center}
\end{table}

The mean Spearman's Rho is 0.5562, this indicates that there is a strong correlation between the rankings produced using the difficulty factor ratings and those that the experts actually gave. It is clear that most experts answer the questions relating to the factors of difficulty in such a way that it is possible to create overall ratings for security components and rankings of attacks that closely match their opinions.

\section{Discussion}
\label{sec:Disc}

The proposed method demonstrates that it is possible to model parts of cyber-security decision makers processes using WAs and OWAs. 

For our first set of experiments we used an EA to search for WA weights that can be used to take experts' ratings of factors that contribute to the difficulty of attacking/evading a security component, and produce overall difficulty ratings for those components. The results showed that for the attack question set it is possible to produce overall ratings that closely match experts' actual rankings. With the evade question set however, much poorer results were obtained. The EA highlighted the factors that contribute the most to the overall rating, specifically questions regarding the availability of tools to help with an attack, the difficulty of achieving the attack unnoticed, the frequency of such attacks being reported and the inherent difficulty of completing an attack. The correlation of experts' answers to these questions and their overall difficulty ratings show that they have moderate/strong correlation, further supporting their importance for this type of assessment. Conversely, questions regarding the maturity of technology, the amount of interaction with data and the complexity of components attracted the least weight and were weakly correlated with the overall ratings. This suggests that these factors were not helpful in this particular exercise, though they may prove to be useful under different circumstances.

The second set of experiments focused on using an EA to find OWA weights that allowed us to aggregate overall ratings of the difficulty of attacking/evading security components to produce rankings of attacks for each expert that are strongly correlated with the actual attack rankings that they provided. This shows that we can extrapolate from ratings of individual components to rankings of complete attacks, lightening the load on experts. 

It was also shown that if the participants are split arbitrarily, OWAs that work well with one group tend to well with the other too. This is a useful result, as it alludes to the possibility of using OWAs to aggregate unseen experts' hop ratings to produce ratings and rankings of unseen attacks for a proposed system. Comprehensive experiments showed that from 30 different starting positions (i.e., seeds) the EA consistently found similar solutions, suggesting that they represent globally good solutions, and that in all 30 runs the EA never placed less than 92\% of the weight on the most difficult component. From this, we can infer that when assessing the difficulty of an attack, the most difficult to attack / evade security component is by far the most important. Simply taking the maximum proves impractical however, as it results in many attacks being given the same rating and ranking.

The final set of experiments combined the processes involved in the first and second sets of experiments. The best WAs and OWA found in previous experiments were used together to produce rankings of attacks from experts' ratings of factors that contribute to the difficulty of a component. The rankings produced, on average, show strong correlation with experts' actual rankings. This indicates that by asking targeted questions about characteristics of security components, specific technical attacks on proposed systems can be rated and ranked in terms of difficulty and salience.
From a machine learning point of view, the proposed methodology where the parameters (i.e. weights) of aggregation operators (both WA and OWA) are optimised using an EA is clearly not the only viable approach. Different optimisation strategies such as simulated annealing could be explored, while the actual aggregation operators could be generalised for example to fuzzy integrals \citep{grabisch2000} where more complex weights captured by fuzzy measures enable a more fine-grained weighting. 

These results raise the possibility of using ratings of security components to produce ratings and rankings for unseen attacks on proposed information systems. This could take the form of a database of ratings of generic security components (we have seen that lack of technical detail does not prevent reasoned assessment) which could be used for \emph{`What if?'} scenarios to give an approximate assessment of the difficulty of attacks when security components are altered or moved. The degree to which this exciting possibility is feasible in practice is a matter of future research.

The major contribution of this paper to cyber-security research lies in its use of exploratory data analysis to demonstrate the feasibility of linking the experts rating of components of attacks to the overall ranking of the difficulty of the attack vectors themselves. In doing so, it both confirms the technical methodology of using WAs and OWAs to perform the aggregation, as well as confirming that the experts' ratings and rankings do have internal consistency with each other --- which is not a \emph{a priori} given, and in some ways can therefore be taken as refutation of our informally posed null hypothesis (that the security experts are not really expert). Hence, not only do these methods provide practical possibilities for systematising the analysis of attack vectors, but they also offer the future possibility of being used as some form of professional competence test (or accreditation) of security expertise. The results of this work also have the potential to reduce both the cost and time required to perform similar estimates through automation. However, we emphasise that these practical benefits will require further research and development. We are not aware of similar approaches to this problem being reported in literature, making direct comparisons impossible. We hope that this paper will initiate many similar studies in the future.

\section{Conclusions}
\label{sec:Conc}

In this work we have presented a method of finding good aggregation operators for creating rankings of technical attacks on a proposed system and ratings of security components, which has the added advantage of also highlighting salient factors and security components. To do this, we have used a data set collected during a decision making exercise at GCHQ in which cyber-security experts performed security assessment tasks on a realistic system proposal. The results showed that using the proposed method it is possible to produce rankings of technical attacks on a system using ratings of security components, ratings of security components from ratings a specific factors of difficulty, and finally, rankings of technical attacks on a system from ratings of specific factors. These outcomes present the possibility of using ratings of generic security components (or their characteristics) in \emph{`What if?'} scenarios to assess the impact of altering/moving components within a system design.

The work has produced important insights that enable the construction of expert security assessment support tools. Such tools have the potential to reduce the time and effort required of experts for assessments, and enable systems designers to produce approximate security assessments before they seek expert advice. These advances will address growing concerns about the capacity of limited expert resources in view of the increasing complexity and growing number of information systems that are under attack from an ever changing set of attacks.

As noted in Section \ref{sec:Disc}, this paper does not advocate the proposed approach of optimising the parameters of the given aggregation operators as a unique solution. Indeed, it is likely that other approaches will achieve similar, and in the case of more complex approaches such as fuzzy integrals, more nuanced results. However, the proposed architecture provides a viable approach which enables the generation of highly useful results and insight in the challenging area of expert-led cyber security system assessment. The same general methodology may be applied to a variety of similar situations in which the modelling of expert opinion, particularly when aggregating sub-criteria, is desired.

Future work in this area will involve efforts to create the generic security component database mentioned previously and applying the techniques shown here to assess the practicality of providing decision support for systems architects in the process of designing new systems. In addition, we are exploring the applicability of the proposed approach to related areas such as `adversary characterisation' in cyber-threat intelligence.

\section{Acknowledgements}
The research reported here has been partially funded by CESG --- the UK Government's National Technical Authority for Information Assurance (IA) and RCUK's Horizon Digital Economy Research Hub Grant, EP/G065802/1. We also thank the anonymous reviewers who have made many insightful and valuable comments which have been incorporated into the text, thereby improving the paper.

\appendix

\section{EA Configuration Experiments}
\label{sec:AppA}

This appendix contains details of the experiments conducted to find ideal EA configurations for the OWA/WA search.

\subsection{EA for use with OWAs}
\label{sec:AppA_OWA}

This subsection contains the experiments conducted to find an EA configuration to be used in the experiments performed in Section \ref{sec:Attack}

Initial experiments focused on discovery of an ideal configuration for the EA, which is a significant problem in itself. In order to create a practical test schedule that could be completed in a reasonable time frame, these experiments were split into two parts: 1) evolutionary operator proportions and 2) generations and population sizes. 

\subsubsection{Part 1 - Evolutionary Operator Proportions}

Table \ref{tab:EASet1} shows the configurations tested in the first part. For all of these experiments 1\% elitism was applied to a population of 250 over 250 generations. The population and generations values have been chosen arbitrarily in order to explore the operator proportions. Appropriate values for population and generations will be examined in later experiments.

\begin{table}[!htb]
\begin{center}
	\caption{EA-OWA: Evolutionary Operator Proportion Experiments}
	{\begin{tabular}{c|c|c|c}
		\hline
		\textbf{Test} & \textbf{Copy}& \textbf{Cross}& \textbf{Mut}\\
		\hline
		1 & 0.00 & 0.20 & 0.79\\
		2 & 0.20 & 0.20 & 0.59\\
		3 & 0.40 & 0.20 & 0.39\\
		4 & 0.60 & 0.20 & 0.19\\
		5 & 0.79 & 0.20 & 0.00\\
		6 & 0.50 & 0.00 & 0.49\\
		7 & 0.30 & 0.40 & 0.29\\
		8 & 0.20 & 0.60 & 0.19\\
		9 & 0.10 & 0.80 & 0.09\\
		10 & 0.00 & 0.99 & 0.00\\
		\hline
	\end{tabular}}
	\label{tab:EASet1}
\end{center}
\end{table}

The best results for these experiments are given in Table \ref{tab:EASet1Results}. The results for all three groupings show that configurations that have a large proportion of mutation and a smaller proportion of crossover work best, and that the proportion of copy has little effect on the results. As the best individuals are automatically copied from one generation to the next via elitism, this may reduce the impact made by the copy operator. Table \ref{tab:EASet1OWA} gives details of the OWAs found in each of the best experiments listed in Table \ref{tab:EASet1Results}. It can be seen that in all cases, the EA is finding OWAs that place significantly more weight on the most difficult to attack/evade security component than the remaining components.

\begin{table}[!htb]
\begin{center}

	\caption{EA-OWA: Evolutionary Operator Proportions - Best Results}
	{\begin{tabular}{c|c c}
		\hline
		\textbf{Test} & \textbf{Mean Sp.} & \textbf{MSE}\\
		\hline
		\multicolumn{3}{c}{\textbf{Even}}\\
		\hline	
		3 & 0.6866 & 0.1425\\		
		\hline
		\multicolumn{3}{c}{\textbf{Odd}}\\
		\hline
		1 & 0.5713 & 0.2543\\
		\hline
		\multicolumn{3}{c}{\textbf{All}}\\
		\hline
		1 & 0.6159 & 0.2069\\
		2 & 0.6159 & 0.2069\\
		\hline
	\end{tabular}}
	\label{tab:EASet1Results}
\end{center}
\end{table}

\begin{table*}[!htb]
\begin{center}
	\caption{EA-OWA: Evolutionary Operator Proportions - Best OWAs}
	{\begin{tabular}{c|c c c c c c c c}
		\hline
		& \multicolumn{8}{c}{\textbf{Best Weights}}\\
		\textbf{Test} & \textbf{1}& \textbf{2}& \textbf{3}& \textbf{4}& \textbf{5}& \textbf{6}& \textbf{7}& \textbf{8}\\
		\hline
		\multicolumn{9}{c}{\textbf{Even}}\\
		\hline	
		3 & 0.8985 & 0.0059 & 0.0311 & 0.0102 & 0.0065 & 0.0258 & 0.0037 & 0.0184\\		
		\hline
		\multicolumn{9}{c}{\textbf{Odd}}\\
		\hline
		1 & 0.7872 & 0.0185 & 0.1242 & 0.0505 & 0.0111 & 0.0032 & 0.0029 & 0.0024\\
		\hline
		\multicolumn{9}{c}{\textbf{All}}\\
		\hline
		1 & 0.9582 & 0.0028 & 0.0242 & 0.0003 & 0.0057 & 0.0029 & 0.0002 & 0.0056\\
		2 & 0.9621 & 0.0017 & 0.0221 & 0.0003 & 0.0045 & 0.0059 & 0.0009 & 0.0025\\
		\hline
	\end{tabular}}
	\label{tab:EASet1OWA}
\end{center}
\end{table*}

\subsubsection{Part 2 - Population and Generations}

In the second part of the EA configuration experiments, the focus switched to the number of generations and population size. The best evolutionary operator proportions from the first part of testing were used in a series of experiments with varying population sizes/numbers of generations. Table \ref{tab:EASet2} provides the population sizes and numbers of generations tested. The best results from these experiments are shown in Table \ref{tab:EASet2Results}. It should be noted that in the first part two configurations were joint best for the \emph{all} group, both were tested in this part. The best result for the \emph{all} group shown in Table \ref{tab:EASet2Results} was achieved with operator proportions from Test 2. In this set of experiments there is much less variation in the results than was seen in the first part of testing; altering the operator proportions had a greater effect on results than altering the population sizes and number of generations. Table \ref{tab:EASet2OWA} gives details of the OWAs found in each of the best experiments listed in Table \ref{tab:EASet2Results}. Again, the EA is finding OWAs that place significantly more weight on the most difficult to attack/evade security component than the remaining components.

\begin{table}[!htb]
\begin{center}
	\caption{EA-OWA: Population and Generations Experiments}
	{\begin{tabular}{c|c|c}
		\hline
		\textbf{Test} & \textbf{Gens}& \textbf{Pop}\\
		\hline
		11 & 50 & 1250\\
		12 & 100 & 625\\
		13 & 200 & 315\\
		14 & 300 & 210\\
		15 & 400 & 155\\
		16 & 500 & 125\\
		\hline  
	\end{tabular}}
	\label{tab:EASet2}
\end{center}
\end{table}

\begin{table}[!htb]
\begin{center}
	\caption{EA-OWA: Population and Generations - Best Results}
	{\begin{tabular}{c|c c}
		\hline
		\textbf{Test} & \textbf{Mean Sp.} & \textbf{MSE}\\
		\hline
		\multicolumn{3}{c}{\textbf{Even}}\\
		\hline	
		12 & 0.6885 & 0.1421\\
		13 & 0.6866 & 0.1421\\		
		\hline
		\multicolumn{3}{c}{\textbf{Odd}}\\
		\hline
		13 & 0.5732 & 0.2519\\
		16 & 0.5733 & 0.2519\\
		\hline
		\multicolumn{3}{c}{\textbf{All}}\\
		\hline
		15 & 0.6165 & 0.2080\\
		\hline
	\end{tabular}}
	\label{tab:EASet2Results}
\end{center}
\end{table}

\begin{table*}[!htb]
\begin{center}
	\caption{EA-OWA: Population and Generations - Best OWAs}
	{\begin{tabular}{c|c c c c c c c c}
		\hline
		& \multicolumn{8}{c}{\textbf{Best Weights}}\\
		\textbf{Test} & \textbf{1}& \textbf{2}& \textbf{3}& \textbf{4}& \textbf{5}& \textbf{6}& \textbf{7}& \textbf{8}\\
		\hline
		\multicolumn{9}{c}{\textbf{Even}}\\
		\hline	
		12 & 0.8907 & 0.0011 & 0.0323 & 0.0120 & 0.0070 & 0.0174 & 0.0035 & 0.0359\\
		13 & 0.8899 & 0.0044 & 0.0297 & 0.0126 & 0.0078 & 0.0242 & 0.0030 & 0.0284\\		
		\hline
		\multicolumn{9}{c}{\textbf{Odd}}\\
		\hline
		13 & 0.7858 & 0.0223 & 0.1285 & 0.0471 & 0.0115 & 0.0031 & 0.0015 & 0.0002\\
		16 & 0.7854 & 0.0246 & 0.1248 & 0.0479 & 0.0117 & 0.0033 & 0.0001 & 0.0021\\
		\hline
		\multicolumn{9}{c}{\textbf{All}}\\
		\hline
		15 & 0.9456 & 0.0022 & 0.0296 & 0.0006 & 0.0068 & 0.0011 & 0.0086 & 0.0055\\
		\hline
	\end{tabular}}
	\label{tab:EASet2OWA}
\end{center}
\end{table*}

\section{EA for use with WAs}
\label{sec:AppA_WA}

This subsection contains the experiments conducted to find an EA configuration to be used in the experiments performed in Section \ref{sec:Comp}

\subsection{Part 1 - Evolutionary Operator Proportions}

In the first set of experiments a series of evolutionary operator proportions are tested. As in the previous experiments, a fixed population of 250 is run over 250 generations and 1\% elitism is applied. Table \ref{tab:EAWASet1} shows details of the configurations tested, the best results for each question set are shown in Table \ref{tab:EAWASet1Results}.

\begin{table}[!htb]
\begin{center}
	\caption{EA-WA: Evolutionary Operator Proportion Experiments}
	{\begin{tabular}{c|c|c|c}
		\hline
		\textbf{Test} & \textbf{Copy}& \textbf{Cross}& \textbf{Mut}\\
		\hline
		1 & 0.00 & 0.20 & 0.79\\
		2 & 0.20 & 0.20 & 0.59\\
		3 & 0.40 & 0.20 & 0.39\\
		4 & 0.60 & 0.20 & 0.19\\
		5 & 0.79 & 0.20 & 0.00\\
		6 & 0.50 & 0.00 & 0.49\\
		7 & 0.30 & 0.40 & 0.29\\
		8 & 0.20 & 0.60 & 0.19\\
		9 & 0.10 & 0.80 & 0.09\\
		10 & 0.00 & 0.99 & 0.00\\
		\hline
	\end{tabular}}
	\label{tab:EAWASet1}
\end{center}
\end{table}

\begin{table}[!htb]
\begin{center}
	\caption{EA-WA: Evolutionary Operator Proportions - Best Results}
	{\begin{tabular}{c|c c}
		\hline
		& \multicolumn{2}{c}{}\\
		\textbf{Test} & \textbf{Mean Sp.} & \textbf{MSE}\\
		\hline
		\multicolumn{3}{c}{\textbf{Attack}}\\
		\hline	
		2 & 0.7936 & 194.0279\\		
		\hline
		\multicolumn{3}{c}{\textbf{Evade}}\\
		\hline
		1 & 0.2270 & 420.2156\\
		\hline
	\end{tabular}}
	\label{tab:EAWASet1Results}
\end{center}
\end{table}

These initial results show that the results are significantly different for each question set. The attack questions produce ratings that are extremely strongly correlated with experts' actual overall difficulty ratings, while the WAs for the evade questions produce ratings that are weakly correlated with experts' actual ratings.


\subsection{Part 2 - Population and Generations}

In the second part of the EA configuration experiments the best evolutionary operator proportions found for each question set in part 1 are used in a series of experiments with varying population sizes and numbers of generations. Table \ref{tab:EAWASet2} provides the configurations tested and Table \ref{tab:EAWASet2Results} shows the best results.

\begin{table}[!htb]
\begin{center}
	\caption{EA-WA: Population and Generations Experiments}
	{\begin{tabular}{c|c|c}
		\hline
		\textbf{Test} & \textbf{Gens}& \textbf{Pop}\\
		\hline
		11 & 50 & 1250\\
		12 & 100 & 625\\
		13 & 200 & 315\\
		14 & 300 & 210\\
		15 & 400 & 155\\
		16 & 500 & 125\\
		\hline
	\end{tabular}}
	\label{tab:EAWASet2}
\end{center}
\end{table}

\begin{table}[!htb]
\begin{center}
	\caption{EA-WA: Population and Generations - Best Results}
	{\begin{tabular}{c|c c}
		\hline
		& \multicolumn{2}{c}{}\\
		\textbf{Test} & \textbf{Mean Sp.} & \textbf{MSE}\\
		\hline
		\multicolumn{3}{c}{\textbf{Attack}}\\
		\hline	
		15 & 0.7931 & 194.0125\\
		\hline
		\multicolumn{3}{c}{\textbf{Evade}}\\
		\hline
		11 - 16 & 0.2270 & 420.2156\\
		\hline
	\end{tabular}}
	\label{tab:EAWASet2Results}
\end{center}
\end{table}

Again, the results for the attack question set are far superior to those for the evade questions.

\section*{\refname}
\bibliography{FuzzySecurity,COSE}
\end{document}